# Effects of Illumination on the Categorization of Shiny Materials

J. Farley Norman
Western Kentucky University

James T. Todd
Ohio State University

Flip Phillips
Skidmore College

Running Head:  Categorization of shiny materials

Correspondence should be addressed to James T. Todd, Department of Psychology, The Ohio State University, Columbus, OH 43210, Phone 614-292-8661, e-mail todd.44@osu.edu.

This research was supported by a grant from the National Science Foundation (BCS-1849418).

**Abstract**

The present research was designed to examine how patterns of illumination influence the perceptual categorization of metal, shiny black, and shiny white materials. The stimuli depicted three possible objects that were illuminated by five possible HDRI light maps, which varied in their overall distributions of illuminant directions and intensities. The surfaces included a low roughness chrome material, a shiny black material, and a shiny white material with both diffuse and specular components. Observers rated each stimulus by adjusting four sliders to indicate their confidence that the depicted material was metal, shiny black, shiny white or something else, and these adjustments were constrained so that the sum of all four settings was always 100%. The results revealed that the metal and shiny black categories are easily confused. For example, metal materials with low intensity light maps or a narrow range of illuminant directions are often judged as shiny black, whereas shiny black materials with high intensity light maps or a wide range of illuminant directions are often judged as metal. In an effort to discover the visual information on which these judgements are based, we measured the specular coverage of each image, and the results were highly correlated with the observers' confidence ratings. We also performed a spherical harmonic analysis on the different light maps in an effort to quantitatively predict how they would bias observers' judgments of metal and shiny black surfaces.

## Effects of Illumination on the Categorization of Shiny Materials

During the past decade, there has been a growing amount of interest in the ability of observers to perceptually identify different types of surface materials. For example, Sharan, Rosenholtz & Adelson (2009, 2014) have shown that observers can rapidly identify material categories such as metal, glass or fabric from briefly presented photographs of objects in natural settings (see also Wiebel, Valsecchi & Gegenfurtner, 2013). Their results reveal that observers can achieve 80% accuracy with presentation times as low as 40 msec, which is much better performance that can currently be achieved by computer vision models of material recognition (Liu et al., 2010; Hu, Bo & Ren, 2010; Sharan et al., 2013).

Although these findings provide clear evidence that observers can identify surface materials, they do not reveal the specific sources of information on which these judgments are based. There is some evidence to suggest that the perception of material properties may involve heuristic processes that can sometimes produce systematic errors. Much of this research has focused on the perception of gloss (Adams, et al., 2018; Doerschner et al., 2010; Marlow & Anderson, 2013, 2015; Mooney & Anderson, 2014; Nishida & Shinya, 1998; Olkkonen & Brainard, 2010, 2011; Pont & te Pas, 2006; Zhang et al., 2015) and the perception of translucency (Fleming & Bülthoff, 2005; Marlow, Kim & Anderson, 2017; Xiao et al, 2014). The results show clearly that observers' judgments of these properties can be influenced by factors that are physically independent of an object's material composition, such as its 3D shape or its pattern of illumination.

Todd & Norman (2018) have recently reported that the pattern of illumination can also influence the perceptual distinction between shiny metal and shiny black dielectric materials such as obsidian. It is important to recognize that the reflection of light from both of these material classes is entirely specular, so that there is only a tiny range of incident angles for each local surface region that will reflect any light toward the point of observation. To better understand the differences between these materials, it is useful to consider how reflectance varies as a function of the incident angle of illumination based on the Fresnel equations. Figure 1 shows the reflectance functions for silver, chrome and obsidian (i.e. volcanic glass). Note that silver reflects almost 100% of the incident illumination at all incident angles, whereas obsidian (or shiny black plastic) reflects close to 0% except at relatively high incident angles. Chrome is roughly midway between those two extremes.

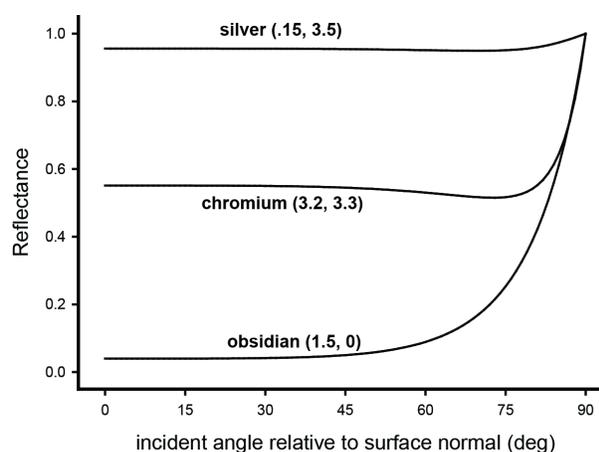

**Figure 1** – Specular reflectance as a function of the incident angle for silver, chromium and obsidian in unpolarized light. The numbers in parentheses show the real and complex coefficients of the index of refraction.

The perceptual distinction between metal and obsidian is similar in some respects to the classic problem of lightness constancy, in that the luminance of any given surface patch is determined by the product of its reflectance and illumination. By analogy, this suggests that metal and shiny black materials

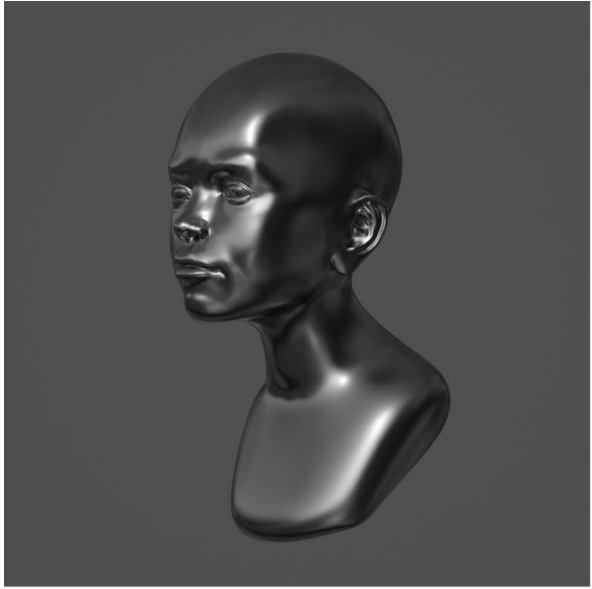
Metal, illumination 1x

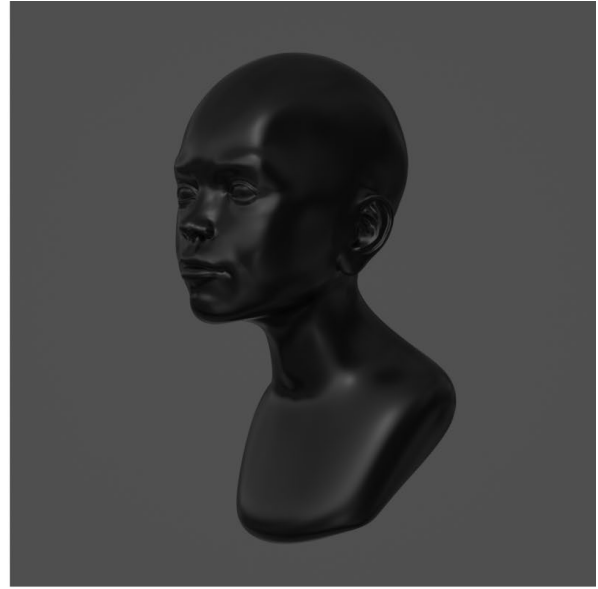
Shiny black, illumination 1x

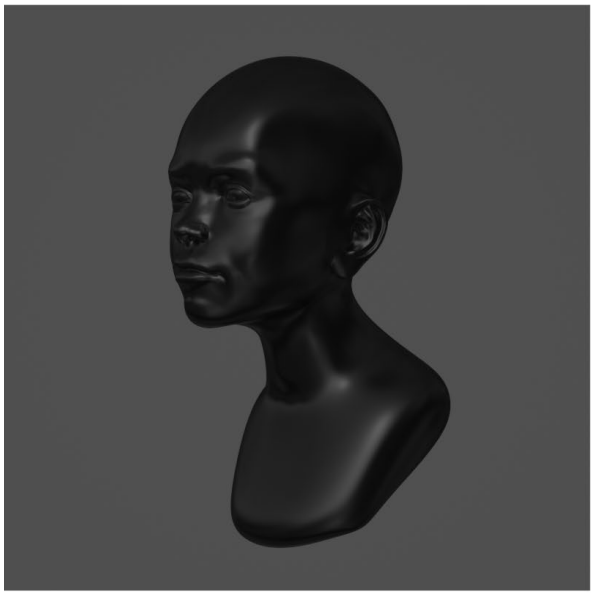
Metal, illumination .2x

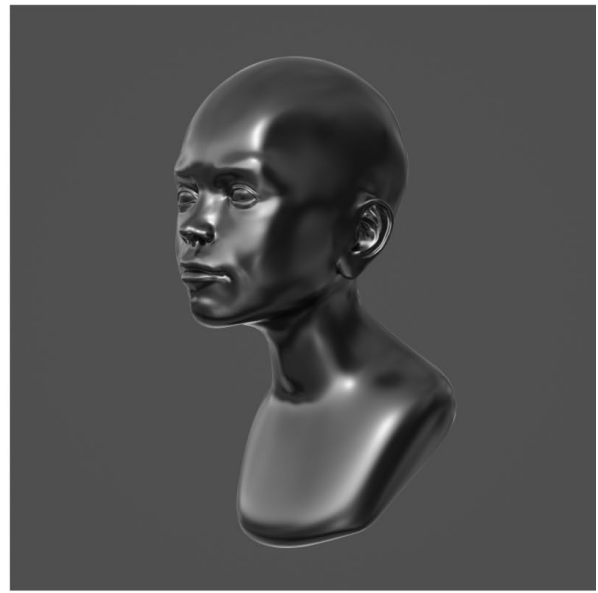
Shiny black, illumination 5x

**Figure 2 –** Four images of a boy's bust made with chromium and shiny black materials. All of the objects are illuminated by the same HDRI light map, although the overall intensity was varied as shown in the caption below each image.

could potentially be confused by selectively manipulating the intensity of illumination. Consider, for example, the two images of a boy's bust in the top row of Figure 2. Both objects are illuminated by an HDRI light probe of an esplanade, and they both have the same magnitude of illumination and the same camera exposure. The one on the left depicts a low roughness chrome material that is perceived as metal. The image on the right depicts a low roughness obsidian material that is perceived as shiny black. It is important to keep in mind that variations in reflectance can be offset by variations in the magnitude of illumination or camera exposure. The bottom left panel of Figure 2 shows a low roughness chrome material with a fivefold reduction in illumination that is perceived as shiny black, and the bottom right panel shows a low roughness obsidian material with a fivefold increase in illumination that is perceived as metal.

It is interesting to note that there is a possible source of information that could potentially be used to perceptually distinguish the metal surface from the shiny black one with high illumination. Because of the Fresnel effect, the peak specular reflections on metal surfaces tend to be in surface regions that face toward the point of observation, whereas the peak specular reflections of shiny black surfaces tend to be in peripheral regions near smooth occlusion boundaries that face away from the point of observation (see Todd & Norman, 2018). With careful inspection, this can be observed in Figure 2. Note that the shiny black object in the lower right panel has a distinct brightening along much of its occlusion contour, which is not present in the metal object in the upper left panel. This effect is rather subtle, however, and does not seem to have much impact on how these objects are perceived.

Another important factor that can influence the perceptual distinction between metal and shiny black materials is the distribution of illuminant directions. Whenever that distribution is relatively broad, the specular reflections on shiny black materials will be sparser than those that occur on metal. This is because shiny black materials have a wide range of incident angles from 0° to 60° for which only a tiny portion of the incident illumination is reflected (see Figure 1). Although this is a potentially useful source of information, a similar sparse pattern of specular reflections can also occur on metal surfaces that are illuminated from a sparse set of directions. Consider the three images of a metal object that are shown in Figure 3. The object in the left panel is illuminated by an HDRI light map of an exhibit hall with a sparse pattern of lights in an otherwise dark environment. It is perceived as a shiny black material. The object depicted in the middle panel is illuminated by a light map of an esplanade, which has an intermediate range of directions. The depicted material in that case is perceived as metal. Finally, the image in the right panel is illuminated by a light map of a snowfield on a cloudy day, which is similar to a Ganzfeld because there is incident light from almost all directions. When that image is shown to human observers they express uncertainty about whether it is metal, shiny white or something else. Note that these perceptual distinctions are based entirely on the pattern of illumination, because all of the depicted objects are composed of the same metal material, and they all have exactly the same 3D shape.

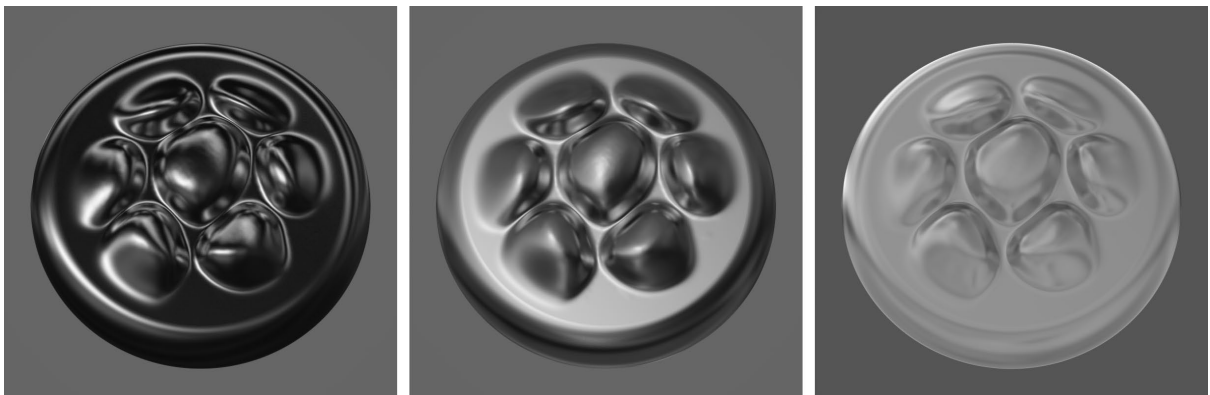

**Figure 3** – Three images of a chromium object illuminated by three different HDRI light maps. The one on the left has a narrow range of illuminant directions and is perceived as shiny black; the one in the middle has an intermediate range and is perceived as metal; and one on the right has a very broad range of directions, and is perceptually ambiguous between metal, shiny white and something else.

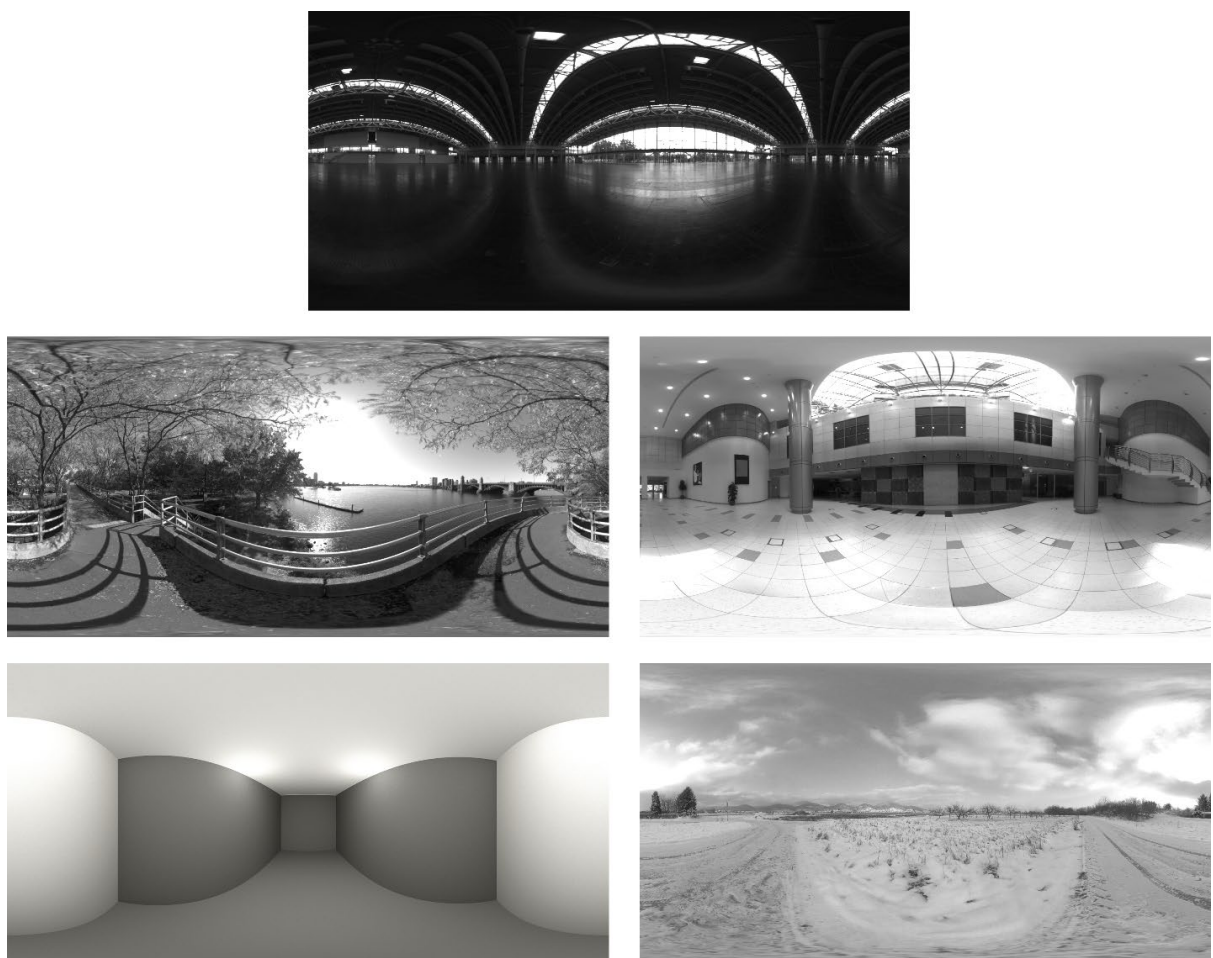

**Figure 4** – Five HDRI light maps used in the present experiment with variable distributions of illuminant directions and intensities.

The research described in the present article was designed to provide a more rigorous exploration of the effects demonstrated in Figures 2 and 3. Observers made material category confidence ratings for objects composed of metal, shiny black or shiny white materials over a wide range of illuminations. The results reveal that the perceptual categorization of these materials is only loosely coupled to the ground truth because observers' judgments are also heavily influenced by the pattern and intensity of illumination.

## Methods

### Material simulations

All of the rendered images presented in this paper were created using the Maxwell Renderer developed by Next Limit Technologies (Madrid, Spain). Maxwell is an unbiased renderer in that it does not use heuristics to speed up rendering times at the cost of physical accuracy. The depicted scenes were illuminated by the five desaturated HDRI light maps shown in Figure 4. They depict an empty exhibit hall, an atrium, an esplanade, an empty white room, and a snowfield on a cloudy day. These light maps were chosen specifically in order to vary the distributions of illuminant directions. For example, the exhibit hall light map has a very sparse set of illuminant directions; the atrium and the esplanade light maps have an intermediate range of illuminant directions; and the white room and snowfield light maps have a broad range of illuminant directions.

In order to appear metallic or shiny, it is important for a material to have a low level of roughness (Todd & Norman, 2018), but there are very few objects in the natural environment that are perfectly smooth with a roughness of zero. Thus, the materials in the present study were modeled using a relatively low roughness of 15. These included a chromium material, whose real and imaginary components of the index of refraction (IOR) were 3.21 and 3.30, respectively. A shiny black material was also included with an IOR of (1.51, 0). Finally, a shiny white material was simulated with a specular reflectance that was identical to the shiny black material, combined with an equal proportion of diffuse reflectance with a roughness of 95 and an IOR of (3, 0). Figure 5 shows examples of these three materials for a single object illuminated by the atrium light map. A more detailed discussion of how the complex IOR influences reflections is provided in the appendix.

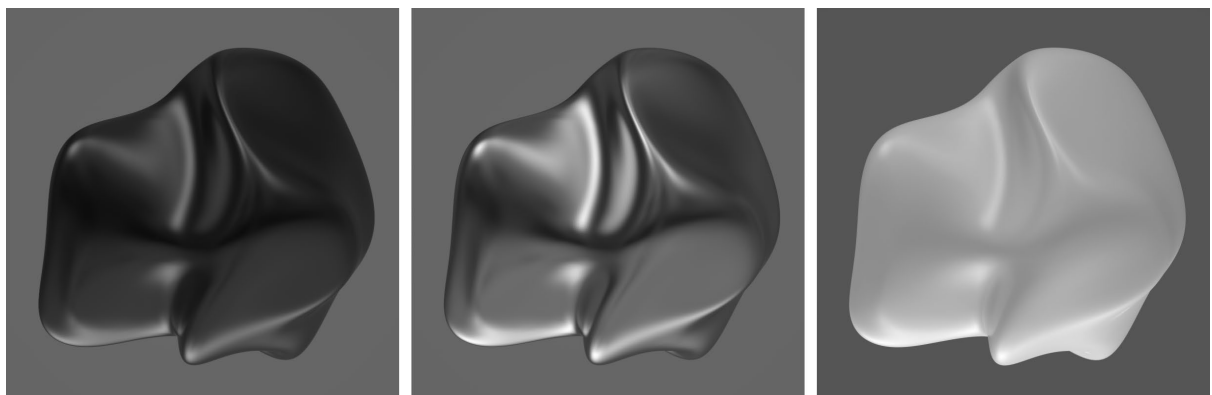

**Figure 5 –** Images of shiny black (left), metal (middle), and shiny white (right) materials, with the same 3D shape and pattern of illumination.

**Apparatus**

The experimental stimulus images were displayed by an Apple Mac Pro computer (Dual Quad-Core processors, with ATI Radeon HD 5770 hardware-accelerated graphics) using an Apple 27-inch LED Cinema Display (2560 x 1440 pixel resolution). The monitor was located at a 100 cm viewing distance. The luminous intensity of the monitor, measured over an area of 25°, had a minimum intensity (for black) of 1 cd/m$^2$ and a maximum intensity (for white) of 136 cd/m$^2$.

**Procedure**

The stimuli depicted three different 3D objects, including the bust of a young boy, a circular disk with seven small bumps, and a randomly deformed sphere (see Figures 2, 3, and 5). The material composition of these objects could be metal, shiny black or shiny white, and they were illuminated by the five possible HDRI light maps shown in Figure 4. The relative intensities of the light maps were adjusted so that the metal objects would all have approximately the same maximum luminance that was just below the maximum image intensity of 255. This was done to ensure that there was no saturation in the depicted specular highlights. These same intensities were employed for the shiny black and shiny white stimuli. However, we also included additional images of metal surfaces for which the base illumination intensity was decreased by a factor of five, and shiny black surfaces for which the base illumination intensity was increased by a factor of five (see Figure 2). We made small downward adjustments to the high illuminations when necessary to avoid saturation of the specular highlights.

The rendered images were globally tone mapped for the Apple monitor into the sRGB 2.1 color space with a D65 white-point and a gamma of 2.2. No other global histogram adjustments (e.g., tint or burn) or local sharpening or contrast enhancement operators were used. Because the intensity of the light maps were adjusted to prevent saturation of the specular highlights and we did not compress the dynamic range of intensities, this likely caused some loss of information at lower intensities that might have been visible on a display device with a higher dynamic range. All of the depicted stimulus objects were presented against a uniform gray background (with an intensity of 100), which was created using an environment map.

On each trial, observers were presented with a single image and were required to categorize the depicted material by adjusting four sliders with a hand-held mouse. Each of the sliders represented a different category labeled metal, shiny black, shiny white or something else, and a digital readout was also provided for each one. Observers were instructed to adjust the sliders to indicate their confidence rating for each of the four possible categories. These confidence ratings were constrained by the program so that the four different ratings would always sum to 100%.

**Observers**

The 75 stimulus images were judged by one of the authors (JFN), and seven other observers who were completely naïve about the purpose of the experiment or how the displays were generated. All observers possessed normal or corrected-to-normal visual acuity. During each experimental session, observers made judgments for all of the 75 stimuli. At the beginning of each session, the details of the response task were explained, and observers were shown real physical examples of metal, shiny black and shiny white materials. All observers participated in two sessions on separate days.

<div align="center">

**Results**

</div>

Figure 6 shows the average confidence rating for each of the four possible response categories collapsed over objects and observers. Each of the individual bar graphs shows the average data for a single light map for all possible combinations of materials and illumination intensities. Let us first consider the results for the shiny white material. Note in Figure 6, that these were categorized as shiny white with a high confidence rating for all five light maps. This finding suggests that the categorization of shiny white materials is only minimally influenced by the pattern of illumination.

There was much more confusion between the metal and shiny black materials, and the pattern of illumination had a much larger influence on the perception of those materials. For example, when images were rendered using the exhibit hall light map with a sparse distribution of illuminant directions, all of the metal objects were categorized as shiny black with a high confidence rating. A quite different pattern of results was obtained with the atrium and esplanade light maps, which had intermediate distributions of illuminant directions: The metal material with a base illumination intensity was rated primarily as metal, and the shiny black material with the base illumination intensity was rated primarily as shiny black. However, when the metal objects were presented with an illumination intensity that was five times lower, they were rated primarily as shiny black, and when the shiny black objects were presented with an illumination intensity that was five times higher, they were rated primarily as metal. These effects are all demonstrated in Figure 2, which was rendered using the esplanade light map.

When the images were rendered with the snowfield or white room light maps, which had broad distributions of illuminant directions, the results were the opposite of those obtained with the exhibit hall. That is to say, there was a general bias to judge all of the shiny black materials as metal. Observers' judgments for the snowfield light map stand out from the others in several respects. First, the primary confusion for the base illumination metal objects was shiny white rather than shiny black. Second, the combined metal and shiny black confidence ratings were lower than in the other conditions, and third, that was the only light map for which the depicted materials were categorized as "something else" with a rating that was significantly above zero. It is important to note that the distribution of illuminant directions for the snowfield light map is close to a Ganzfeld. Images of objects that are illuminated in that manner can look a bit weird (e.g., see right panel of Figure 3), because they have so little contrast. These findings highlight an interesting problem of how observers can distinguish between diffuse and specular reflections when they both have the same color. The problem arises because specular reflections from a Ganzfeld are quite similar to diffuse reflections, though they may sometimes be distinguishable due to specular inter-reflections in concave regions (e.g. see right panel of Figure 3).

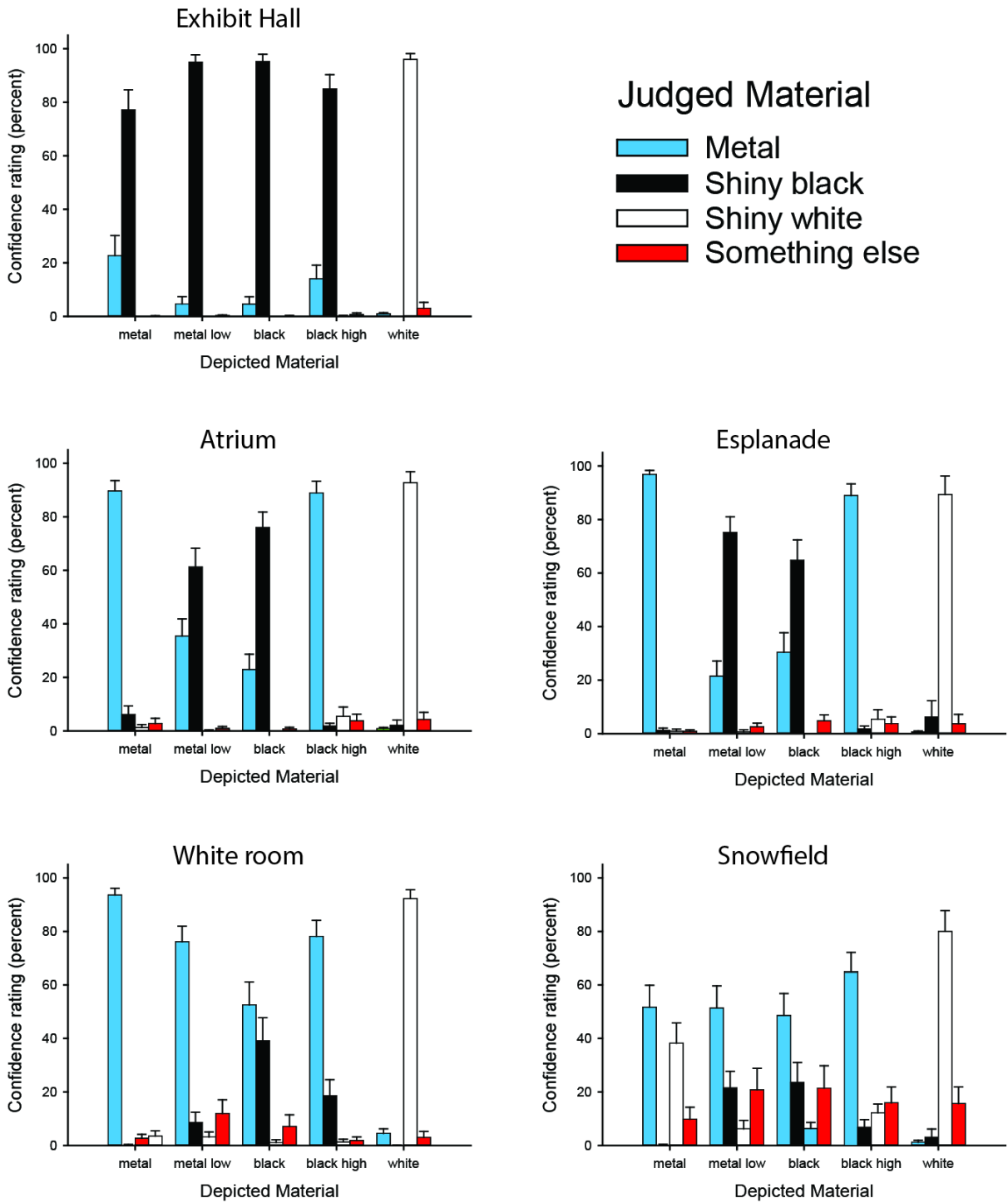

**Figure 6** – The average confidence rating for each of the four possible response categories for all of the different illumination and material conditions, collapsed over objects and observers. Error bars show the standard error of the mean for each condition.

In order to provide a simple quantitative measure of how each light map biased the observers' judgments, we calculated the average metal confidence rating for all of the metal and shiny black materials, as well as the average shiny black confidence rating. The ratio of these two averages provides a bias index for any particular light map. For the five light maps used in the present study, the bias index was 0.13 for the exhibit hall, 1.63 for the atrium, 1.66 for the esplanade, 4.16 for the white room, and 4.51 for the snowfield. These values indicate that the exhibit hall produces a strong bias to perceive purely specular surfaces as shiny black. The atrium and the esplanade produce small biases to perceive purely specular surfaces as metal, whereas the white room and snowfield produce much stronger biases to perceive those surfaces as metal.

Marlow and Anderson (2013) have argued that visual information for the perception of gloss has three component dimensions. One of these called specular contrast refers to differences between the diffuse and specular components of reflection. Note for example that images of shiny black materials typically have much higher specular contrast than those that depict shiny white materials. The second component called specular sharpness refers to the steepness of the luminance gradients along the edges of highlights. The third component of their model is called specular coverage, and it refers to the proportion of an object's surface that is covered by specular reflections. This is the component that is most affected by the pattern of illumination, and we suspect it may be an important source of information for distinguishing different types of shiny materials, such as metal or obsidian.

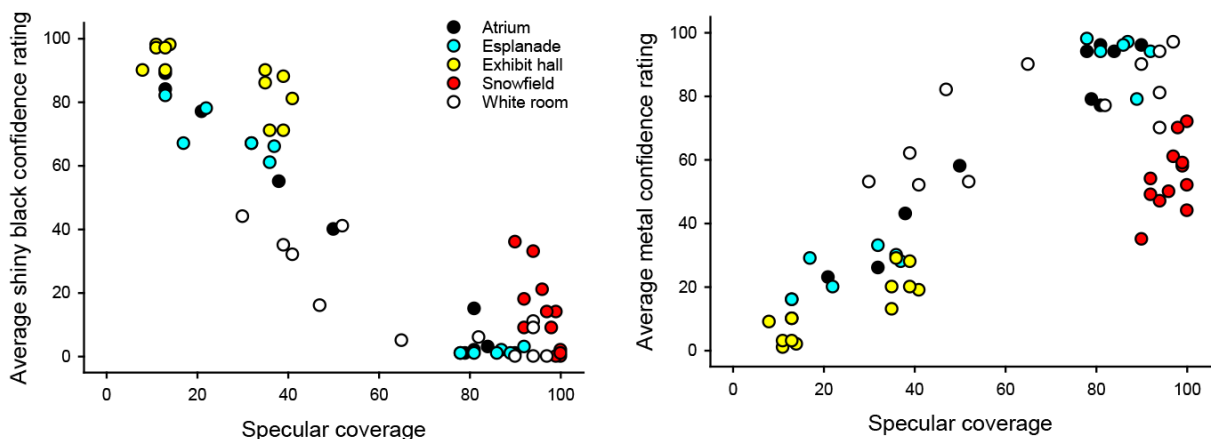

**Figure 7** – The metal and shiny black confidence ratings plotted as a function of specular coverage for each of the five light maps used in the present experiment.

In order to test that hypothesis, we measured the specular coverage for each of the shiny black and metal stimulus images used in the experiment. The shiny white stimuli were excluded from this analysis because they contained both diffuse and specular reflections. Our measure of coverage involved setting a threshold intensity value, and counting the number of pixels with an intensity above that threshold, excluding the background. After some trial and error, we found that a threshold of 50 produced the best fits to the empirical data. The left panel of Figure 7 shows the shiny black confidence ratings as a function of specular coverage, and a similar plot for the metal ratings is shown in the right panel. For the shiny black judgments, there was a strong linear correlation with specular coverage ($R^2 = 0.84$). This relation was more complex for the metal judgments, producing an $R^2$ of 0.69. The outliers in that case included all the stimulus objects illuminated by the snowfield light map. These all had coverage values in excess of 90%, yet the average metal confidence rating in those conditions was only 54%.

We also performed a similar analysis using the mean intensity of the images rather than specular coverage. Although these measures covary to some extent, the coverage measure counts all pixels with an intensity above 50 as equal, whereas the mean intensity weights the brighter pixels more heavily. The linear correlations of mean intensity with the shiny black and metal confidence ratings produced $R^2$ values of 0.61 and 0.50, respectively. Thus, the mean intensity accounts for 21% less variance than the coverage measure, This is consistent with the original hypothesis of Todd & Norman (2018) that specular coverage is a primary factor for distinguishing metal from shiny black materials.

## Discussion

It is important to keep in mind that for any local neighborhood of a purely specular surface, there is only a tiny range of incident angles that will reflect any light toward the point of observation. Thus, the overall specular coverage of these materials is significantly influenced by the distribution of illuminant directions as well. Todd and Norman (2018) proposed a relatively simple rule for the categorization of metal, shiny black and shiny white materials: Shiny white (or colored) materials are perceived when diffuse and specular reflections are additively combined. Shiny black materials are perceived when there are no diffuse reflections, and specular highlights are confined to a relatively small proportion of the visible surface. Metal is perceived when there are no diffuse reflections, and specular reflections are spread out over much of the visible surface. The present experiment was designed specifically to test this hypothesis by systematically varying the patterns of illumination in order to produce varying degrees of specular coverage. The results reveal that the pattern of illumination can influence the categorization of shiny materials almost as much as variations in the actual depicted material.

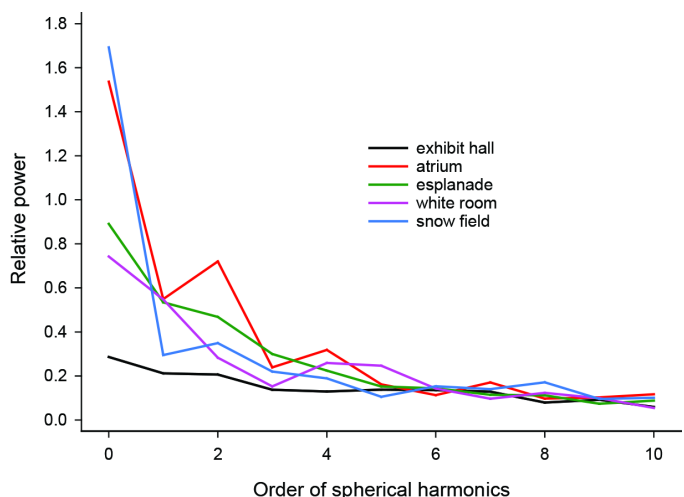

**Figure 8 –** The relative power weightings of the different components of a spherical harmonic analysis (up to the 10th order) for each of the five light maps used in the present experiment.

**The statistical structure of illumination.**

The light maps employed in the present experiment were selected based on a subjective evaluation of the overall distribution of illuminant directions, but it would be useful if these distributions could be quantified in some way. One possible method to achieve this is to decompose the spherical function of the illumination environment by the sum of its spherical harmonics, which is analogous to a Fourier analysis of planar patterns. The power of the spherical harmonics at each order characterizes the angular distributions of the illumination at that order. The zero order component of a spherical harmonic series has a single coefficient that represents the intensity of spherically diffuse ambient light. The first order has three coefficients, each of which represents a light dipole with a source in one direction and a sink in the opposite direction. The relative directions of the three dipoles are all orthogonal to one another. Each subsequent order in the series adds two additional coefficients that represent larger and larger groupings of sources and sinks at finer and finer scales.

Figure 8 shows the relative power of the different components of a spherical harmonic series (up to the 10[th] order) for each of the five light maps used in the present experiment. Each point on these curves represents the root mean squared power for all the coefficients at a given order. Note that the energy drops off quickly with increasing order and that it mostly levels out near zero by the 7[th] order. Zhang et al (2019) have described two possible metrics for describing these distributions: A diffuseness metric (Xia, Pont, & Heynderickx, 2017) is defined as the ratio of the power of the first order harmonic relative to the power of the zero order; and a brilliance metric is defined as the ratio between the sum of the higher harmonics greater than or equal to the third order, relative to the sum of all orders. The values of these metrics (computed to the 30[th] order for brilliance) for each of the five light maps used in the present study are shown in Table 1, together with the bias index for each map that was computed from the observers' confidence ratings. This table also shows that there is a relatively small correlation between each of these metrics and the bias index, resulting in $R^2$ values of 0.18 and 0.40, respectively.

| | | bias index | diffuseness | brilliance | diffuseness2 |
|---|---|---|---|---|---|
| exhibit hall | | 0.13 | 0.74 | 0.79 | 0.72 |
| atrium | | 1.63 | 0.36 | 0.46 | 0.47 |
| esplanade | | 1.66 | 0.60 | 0.55 | 0.53 |
| white room | | 4.16 | 0.74 | 0.53 | 0.38 |
| snowfield | | 4.51 | 0.17 | 0.51 | 0.21 |
| | | | | | |
| | | | $R^2 = 0.18$ | $R^2 = 0.40$ | $R^2 = 0.88$ |

**Table 1** – The bias index for each light map used in the present experiment, and three possible measures of the spherical harmonic series for each map. The bottom row shows the correlations ($R^2$) between each measure and the bias index.

A closer examination of Figure 8 reveals that most of the variance among these particular light maps occurs at orders zero and two. Could the relation between those orders provide a better account of the observers' biases? The fourth column in Table 1 shows another possible metric labeled diffuseness2. Whereas diffuseness is the first order component divided by the zero order, diffuseness2 is the second order component divided by the zero order. Note that the correlation of that measure with the bias index produced an $R^2$ of 0.88, more than twice the values obtained for the brilliance or diffuseness metrics. Because we have no theoretical explanation to justify this particular metric, it is possible that its high correlation with the bias index could be an accidental property of these particular light maps. Nevertheless, among the wide variety of measures we have considered, this is the only one that provides a good fit to the empirical data.

Todd and Norman (2018) have recently demonstrated that it is possible to alter the apparent material of an object between metal and shiny white by selectively filtering the pattern of illumination. In order to follow up on their observations, we created a new set of light maps by filtering some of the ones shown in Figure 4, each of which had a spatial resolution of 4800 X 2400 pixels. Four of these new maps are shown in Figure 9. The ones in the right column were created in HDRshop using an 800 pixel wide Gaussian blur filter on the images of the atrium (top) and esplanade (bottom). This produces low pass filtered images that only contain the lower frequency components of the original light maps shown in

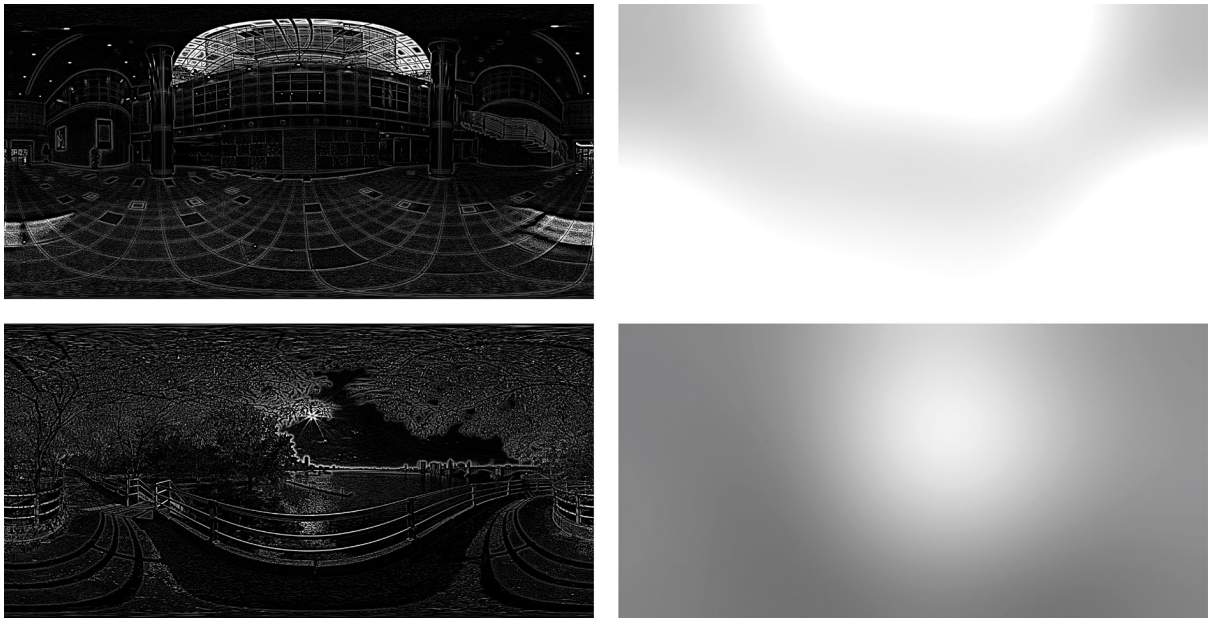

**Figure 9 –** High pass (left) and low pass (right) versions of the atrium (top) and esplanade (bottom) light maps from Figure 4

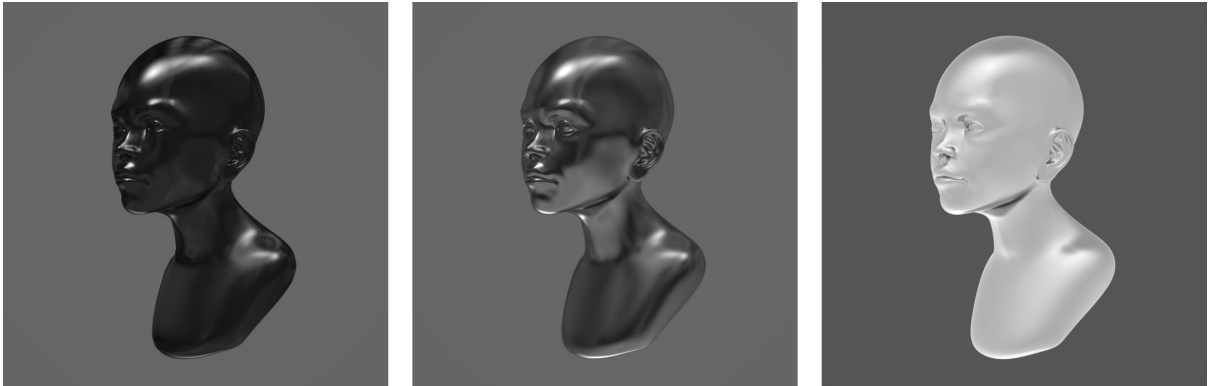

**Figure 10 –** Images of a boy's bust made of polished chrome illuminated using the atrium light map (center) a high pass filtered version of that map (left) and a low pass filtered version (right).

Figure 4. The images in the left column of Figure 9 were created using a 10 pixel wide Gaussian blur filter on the images of the atrium (top) and esplanade (bottom), and then subtracting the blurred images from the originals. This produces high pass filtered images that only contain the higher frequency components of the original light maps shown in Figure 4.

Figure 10 shows three images of a boy's bust made of polished chrome with different patterns of illumination. The image in the middle panel was created with the atrium light map shown in Figure 4. The one on the right was created using the low pass filtered version of the atrium, and the one on the left was created using the high pass filtered version. Note how these variations in illumination have a huge effect on the appearance of the surface materials. The one created with the original atrium light map appears as metal. The one created with the low pass version appears as shiny white (see also Todd & Norman, 2018), and the one created with the high pass version is perceived as shiny black. It is important to point out that the relative intensity of the three light maps were adjusted so that the maximum luminance would be the same in all three images. It is also important to note, however, that the images all have different amounts of specular coverage. The coverage measures from left to right are 17%, 81% and 99%, respectively.

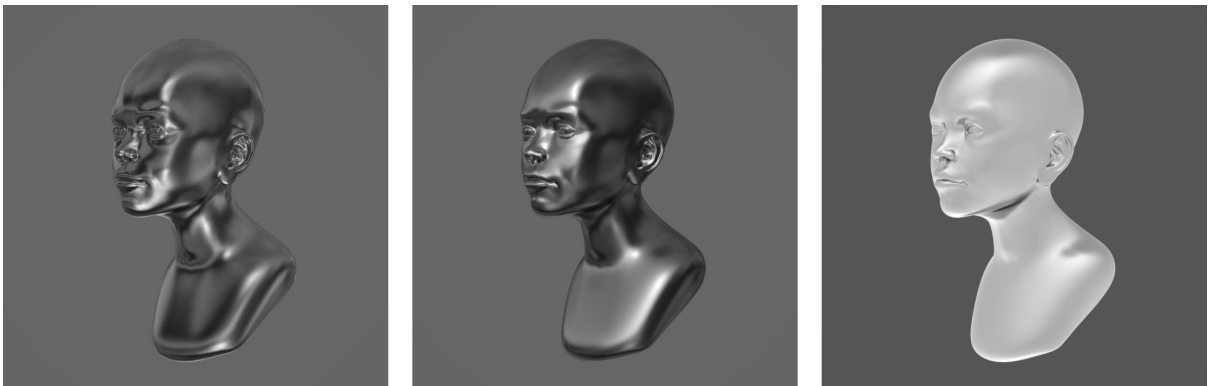

**Figure 11 –** Images of a boy's bust made of polished chrome illuminated using the esplanade light map (center) a high pass filtered version of that map (left) and a low pass filtered version (right).

These examples provide some anecdotal evidence that the higher frequency components of the illumination field are a possible factor that biases metallic surfaces to appear shiny black, shiny white or metallic, but a closer examination reveals that is an oversimplification. Consider the three images in Figure 11 of a boy's bust made of polished chrome illuminated with different versions of the esplanade light map (see Figures 4 and 9). As in Figure 10, when the object is illuminated by the unfiltered light map (middle panel), it appears metallic (coverage = 79%), and when it is illuminated by the low pass version (right panel), it appears as shiny white (coverage = 99%). However, when it is illuminated by a high pass version of the esplanade (left panel), it appears arguably even more metallic (coverage = 85%) than the image produced with the unfiltered original.

The high pass versions of the atrium and the esplanade were created using exactly the same filtering process, but they have quite different effects on the perceptual appearance of the depicted chrome material. It is interesting to note in the left column of Figure 9, that the high frequency energy in the atrium light map is primarily localized in a limited arc of directions near the top, whereas the high frequency energy in the esplanade light map is more broadly distributed over a wider range of directions. If the distinctions between metal, shiny black and shiny white materials are based on specular coverage as originally suggested by Todd and Norman (2018), then it is the distribution of illumination directions rather than spatial frequency per se that is the most important aspect of the illumination field for influencing the appearance of shiny materials.

It turns out that these distributional differences are also captured by the diffuseness2 metric. The value obtained for the high pass version of the atrium light map is 0.90, nearly double the value of 0.49 obtained for the unfiltered version. In contrast, the value obtained for the high pass version of the esplanade light map is only 0.48, which is slightly less than the value of 0.53 that was obtained for the unfiltered version. If high values of the diffuseness2 metric bias observers to perceive specular materials as shiny black rather than metal, this could explain why the high pass atrium light map produces images that appear shiny black (see left panel of Figure 10), whereas the high pass esplanade light map produces images that appear as metal (see left panel of Figure 11).

**Effects of color, roughness and background surfaces**

**Color –** There are several other issues that deserve to be considered involving the generality of these findings. It is important to keep in mind that all of the stimuli in this experiment were achromatic, but that is not typically the case for real objects in the natural environment. For example, consider the brown dielectric material depicted in the left panel of Figure 12. This type of material has two distinct types of reflection: Part of the illumination is transmitted a very short distance into the material, where some of it is scattered back toward the point of observation. This is referred to as diffuse reflection, and its color is determined primarily by the material. Another part of the illumination reflects directly off the object's surface. This is referred to as specular reflection, and its color is determined exclusively by the color of the illumination. The presence of these two different components in combination provides a potentially powerful source of information to indicate that one is looking at a shiny dielectric material. Note that shiny black materials constitute a special case, where that information is unavailable, and the identity of a material as a dielectric must be determined by other means.

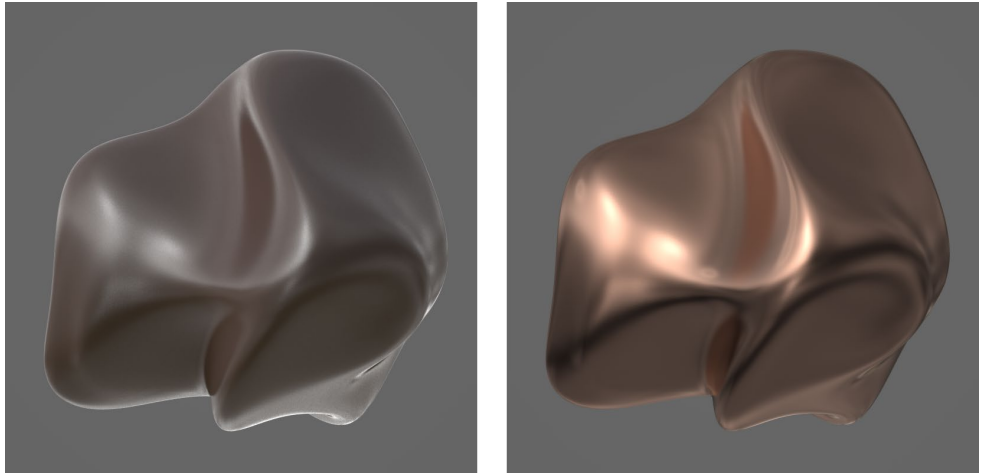

**Figure 12 –** Images of a brown dielectric material (left) and a copper metal material (right)

The right panel of Figure 12 shows a metallic (copper) material with the same base color as the one in the left panel. For metals, 100% of the transmitted light is absorbed, so there are no diffuse reflections. However, unlike dielectrics, the color of the material can influence the colors of its specular reflections. That is why the color of the object in the right panel appears as copper, even though the illumination is completely desaturated. This likely provides useful information for the identification of colored metals, such as copper or gold, but there are other common metals such as silver, aluminum or chrome whose colors are mostly desaturated. The identification of those materials as metal must depend on other factors, like specular coverage.

**Roughness –** In a previous experiment by Todd & Norman (2018) we examined how the appearance of metal and shininess of chrome objects is influenced by variations of surface roughness. When the illumination had a broad range of directions, observers' metal and shininess ratings dropped rapidly with increasing roughness. For roughness values of 60 or higher, the material no longer appeared metallic or shiny, and was perceived instead as matte. Figure 13 shows images of a chrome material (left column) and a purely specular black dielectric material (right column), both with a roughness of 60. Note that the appearance of metal and shiny black have been completely eliminated. Both materials appear matte, although the metal one appears lighter than the black one. The objects in the top row were illuminated using the exhibit hall light map, whereas the ones in the bottom row were illuminated by the esplanade light map. Note that there were significant differences in the metal and shiny black confidence ratings for these light maps when the materials were depicted with low roughness (see Figure 7), but that effect appears to be eliminated when the objects are depicted with high roughness.

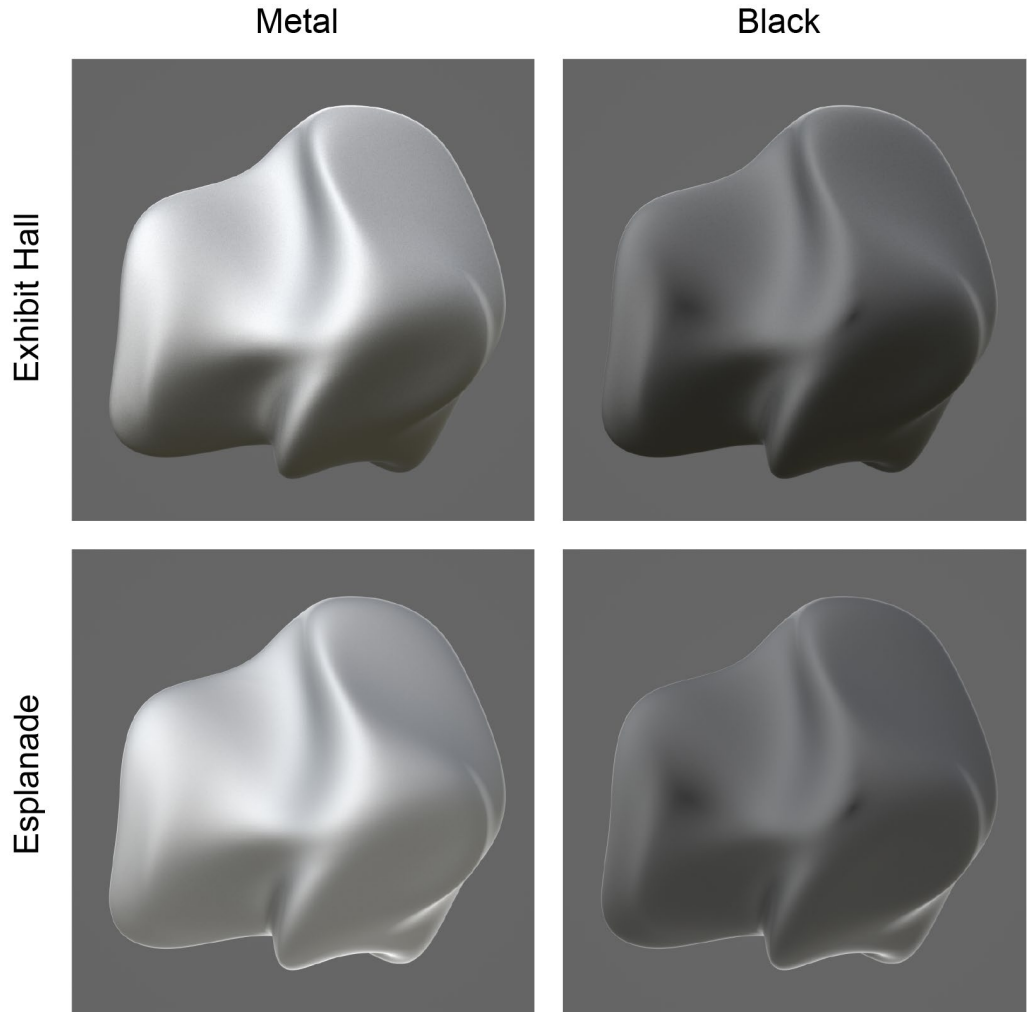

**Figure 13** – Images of a rough metal material (left) and a rough shiny black material (right) illuminated by the exhibit hall and esplanade light maps.

**Background surfaces** – Adams et al (2018) have recently demonstrated that constancy of gloss perception over variations of tone mapping is significantly improved if depicted objects are presented against a background of a natural scene, as opposed to a neutral gray background as in the present experiment. We were curious if contextual information about the lighting might also improve the categorization of metal and shiny black materials, so we created a set of images of a polished metal deformed sphere with the exhibit hall, atrium and snowfield light maps, in which the same maps also provided a background scene. The resulting images are shown in Figure 14. To our eyes, the one on the left appears as shiny black, the one in the middle appears as metal, and the one on the right is ambiguous between metal and shiny white. These examples suggest that scene context may not noticeably improve the material constancy of purely specular surfaces. However, it is possible that the visual structure of a surrounding scene might be more informative for material constancy if the observer is embedded in the scene, rather than viewing it in a picture. That is an issue that will remain to be addressed in future research.

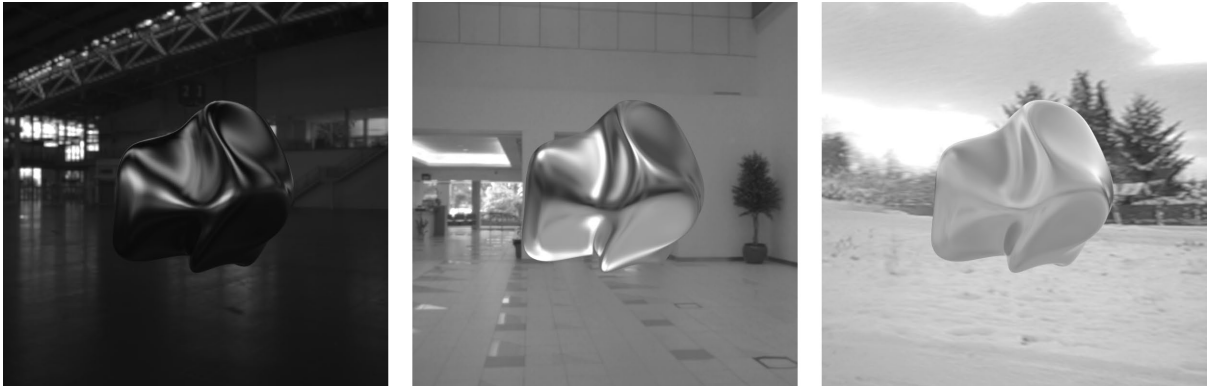

**Figure 14** – Three images of a distorted sphere illuminated by the exhibit hall (left), the atrium (middle) and the snowfield (right), with a visible background to provide more contextual information about each scene.

## Appendix

### The parameterization of surface reflectance

The manner in which a material interacts with light is determined by its index of refraction, which is a complex number (n - ki). The imaginary coefficient (k) is also known as the extinction coefficient. For most of the materials encountered in nature (i.e., dielectrics), the value of k is vanishingly small, but that is not the case for metals. To better appreciate how n and k can influence reflections on metal surfaces it is useful to consider the three images shown in Figure A1. The left panel of this figure depicts an aluminum material with an IOR of (1.2, 7.0), which is perceived as a shiny metal. The image in the middle panel was generated with exactly the same parameters as the one on the left, except that the value of k was set to zero. This changes the appearance to a black dielectric material. However, by raising the value of n to 50, and keeping k at zero, the appearance of a shiny metal can be restored, as shown in the right panel. The pattern of shading in that case is similar (but not identical) to the image in the left panel.

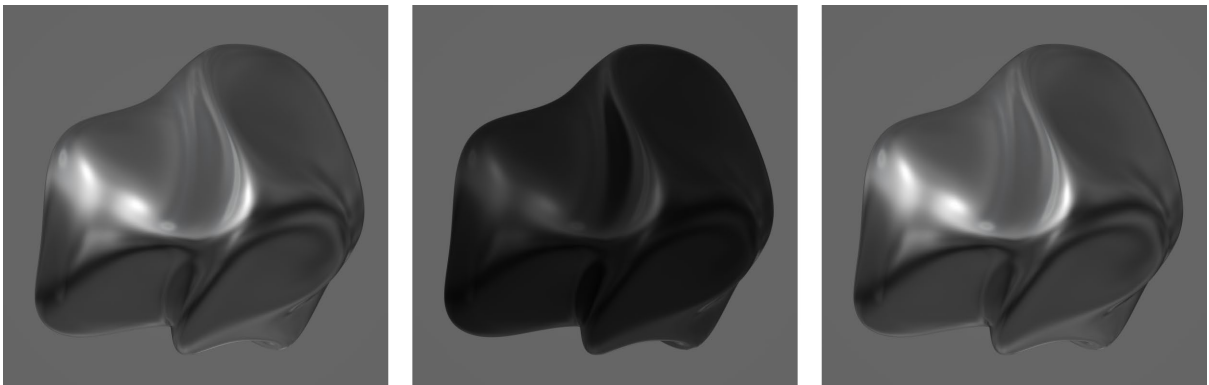

**Figure A1** – Three images of a distorted sphere with different complex IORs. From left to right the IORs are (1.2, 7.0), (1.2, 0) and (50, 0).

The complex IOR is used by the Fresnel equations to compute the surface reflectance at all possible incident angles and directions of polarization. Some renderers like Maxwell, Renderman or Arnold allow the user to input values of n and k to characterize a material. This is the best way to achieve physical accuracy, because the appropriate values can be obtained from a material handbook or at the web site https://refractiveindex.info/. Other renderers employ a different approach. By specifying the reflectance and color at incident angles of 0° and 90° (also referred to as facing and grazing angles), it is possible to compute the values of n and k from that information. This is just a reparameterization of the same space, which some users find more intuitive because it focuses on the end result of the computation.

Still other renderers only allow the user to specify the real coefficient (n) of the IOR, or a facing angle reflectance, and the value of k is assumed to be zero. This approach allows the program to use an approximation of the Fresnel equations, which can greatly reduce the required computations. Metals can only be simulated in these renderers by using an unnaturally high value of n, as shown in the right panel of Figure A1. Another variation of this approach is to incorporate a metalness parameter. Although the term metalness does not occur in physics, it is used in renderers to make the user interface more intuitive. The metalness parameter acts as a switch between metal and dielectric materials. When set to zero, a low value of n is used that is typical of dielectric materials, and the color of the specular reflections is exclusively determined by the color of the illumination. When set to one, a high value of n is used, and the material color is allowed to influence the specular reflections. The IOR parameter in these renderers may only be used for refractions.